\documentclass[10pt,twocolumn,letterpaper]{article}

\usepackage[pagenumbers]{iccv} 

\usepackage[ruled,linesnumbered]{algorithm2e}
\usepackage{bbm}
\usepackage{multirow}

\usepackage{array}
\newcolumntype{C}[1]{>{\centering\arraybackslash}p{#1}}

\definecolor{iccvblue}{rgb}{0.21,0.49,0.74}
\usepackage[pagebackref,breaklinks,colorlinks,allcolors=iccvblue]{hyperref}

\title{MCBLT: Multi-Camera Multi-Object 3D Tracking in Long Videos}

\author{Yizhou Wang$^{1,*}$, Tim Meinhardt$^{1,*}$, Orcun Cetintas$^{1,2}$, Cheng-Yen Yang$^{1,3}$, Sameer S. Pusegaonkar$^{1}$, \\
Benjamin Missaoui$^{1}$, Sujit Biswas$^{1}$, Zheng Tang$^{1}$, Laura Leal-Taixé$^{1}$ \vspace{1mm}\\
$^{1}$ NVIDIA, $^{2}$ Technical University of Munich, $^{3}$ University of Washington\\
{\small $^*$ Equal contribution}\\
{\tt\small \{yizwang, tmeinhardt\}@nvidia.com}
}

\begin{document}
\maketitle
\begin{abstract}
Object perception from multi-view cameras is crucial for intelligent systems, particularly in indoor environments, \eg, warehouses, retail stores, and hospitals. Most traditional multi-target multi-camera (MTMC) detection and tracking methods rely on 2D object detection, single-view multi-object tracking (MOT), and cross-view re-identification (ReID) techniques, without properly handling important 3D information by multi-view image aggregation. In this paper, we propose a 3D object detection and tracking framework, named MCBLT, which first aggregates multi-view images with necessary camera calibration parameters to obtain 3D object detections in bird's-eye view (BEV). Then, we introduce hierarchical graph neural networks (GNNs) to track these 3D detections in BEV for MTMC tracking results. Unlike existing methods, MCBLT has impressive generalizability across different scenes and diverse camera settings, with exceptional capability for long-term association handling. As a result, our proposed MCBLT establishes a new state-of-the-art on the AICity'24 dataset with $81.22$ HOTA, and on the WildTrack dataset with $95.6$ IDF1. 
\end{abstract}    
\section{Introduction}
\label{sec:intro}

Detecting and tracking objects across multiple cameras is a crucial problem for 3D environment understanding, particularly in retail or warehouse settings, for various applications, including inventory management, security surveillance, or customer behavior analysis. 
In these use cases, a typical multi-camera system involves numerous cameras with diverse viewing angles and fields of view (FoV). While some cameras usually have overlapping FoVs, others may not share any common visual space. 
However, this task presents significant challenges such as occlusions, varying lighting conditions, and the need to maintain consistent object identification across different camera views. Moreover, the integration of 3D spatial information from multiple 2D views requires sophisticated algorithms to handle camera calibration errors and perspective distortions. Addressing these issues is crucial for developing robust multi-camera detection and tracking systems that can reliably operate in complex environments.


\begin{figure}[t]
  \centering
   \includegraphics[width=\linewidth]{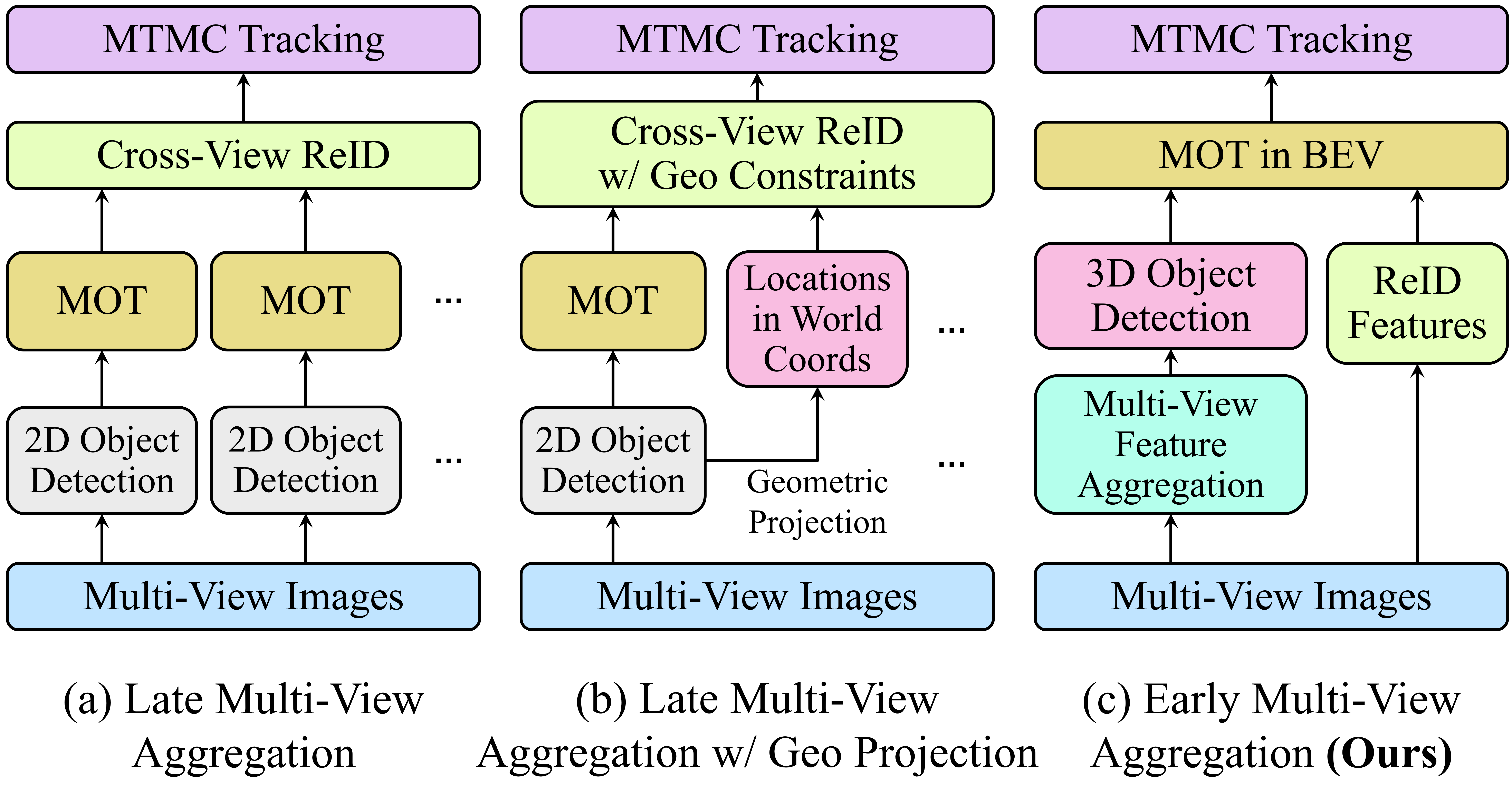}
   \vspace*{-0.6cm}
   \caption{Comparison among three types of MTMC tracking methods. 
   (a) conducts 2D detection separately and associates objects among different views by appearance-based ReID; (b) considers geometric constraints as well besides appearance for cross-view association; (c) achieves multi-view association in early stage by feature-level aggregation.
   }
   \vspace*{-1.5em}
   \label{fig:mtmc_comp}
\end{figure}

Existing MTMC tracking methods can be divided into three categories: (i)~late multi-view aggregation, (ii)~late multi-view aggregation with geometric projection, and (iii)~early multi-view aggregation, as shown in~\cref{fig:mtmc_comp}.
Late multi-view aggregation pipelines detect objects in each camera view as 2D bounding boxes, with some methods tracking these 2D detections separately before associating them across cameras using re-identification (ReID) embeddings and spatial constraints. Late multi-view aggregation pipelines with geometric projection pipelines~\cite{xu2016multi,xu2017cross,cheng2023rest} further perform additional projections using camera calibration matrices to achieve spatial association for each camera view into a global coordinate system. 
Recently, ~\cite{teepe2024earlybird} demonstrated the possibility of early aggregation before any perception steps in each camera view. This method first fuses multi-view image information into a unified 3D space and directly conducts both detection and tracking in this 3D space, which can significantly improve detection quality and avoid association errors due to unreliable spatial alignment among different camera views.
However, its detection network is designed for a fixed multi-camera scene, so it is not flexible for different environments, different numbers of cameras, or different camera placements. Therefore, it does not have good scalability for large scenes with a large number of cameras.
Furthermore, the tracking of~\cite{teepe2024earlybird} merely relies on a heuristic  Kalman filter, which is prone to drifting and struggles with long-term occlusion handling.

In this work, we propose \textbf{MCBLT}, a \textit{robust} and \textit{generalizable} method for \textbf{M}ulti-\textbf{C}amera \textbf{B}ird's-eye-view \textbf{L}ong-term \textbf{T}racking in various complex multi-camera indoor/outdoor public environments. Our proposed approach leverages a unified BEV representation and spatio-temporal transformers to directly infer 3D object bounding boxes. Then, the well-localized 3D detections are backprojected to 2D camera images and matched with 2D detections for optimal 2D appearance feature extraction. This allows us to benefit from both highly accurate 2D and 3D detections for ReID and long-term tracking.
For the temporal association of detections, we introduce the first 3D tracking-by-detection approach utilizing hierarchical graph neural networks (GNNs)~\cite{cetintas2023unifying}.
The lightweight hierarchical structure allows us to learn a GNN tracking model that directly predicts associations for up to thousands of frames.
Compared to prior 2D GNN tracking solutions~\cite{cetintas2023unifying}, we scale the ability to bridge occlusion gaps by an order of magnitude.  
Our model consists of a hierarchy of GNNs that associate 3D detections via geometrical and multi-view ReID embedding features.
To track objects across very long sequences, we introduce a novel global tracking layer that replaces \textit{handcrafted heuristics} required to run the GNN in a sliding window fashion~\cite{cetintas2023unifying} with \textit{a model layer} that operates globally over the video sequence and matches incoming objects with past tracks.
The global layer requires no additional training and significantly boosts occlusion handling and overall tracking performance.

We evaluate MCBLT on a large-scale synthetic indoor MTMC dataset AICity'24~\cite{wang20248th} and a real-world outdoor MTMC dataset WildTrack~\cite{chavdarova2018wildtrack}. MCBLT achieves SOTA on both datasets, with $81.22$ HOTA on AICity'24, and $95.6$ IDF1 on WildTrack.

Overall, our contributions can be summarized as follows:
\begin{itemize}
    \item The first MTMC method efficiently performs early multi-view image aggregation for 3D perception.
    \item A ReID feature extraction method for MTMC tracking based on a 2D-3D detection association algorithm.
    \item The first hierarchical GNN-based 3D multi-object tracking method and a novel global tracking block that unlocks long-term associations across thousands of frames.
    \item We achieve state-of-the-art performance on both the AICity'24 and WildTrack datasets.
\end{itemize}

\section{Related Works}
\label{sec:related}

\paragraph{Multi-View Object Detection}
Object detection from multi-view images is an essential technique for understanding 3D geometric information and handling occlusion. With proper synchronization and calibration, images from multiple viewing angles can be integrated into the same space to accurately learn 3D geometry, \eg, 3D locations, dimensions, and orientations. 
To begin with, researchers develop multi-view object detection methods based on some classical approaches, \eg, conditional random field (CRF)~\cite{baque2017deep,roig2011conditional}, probabilistic modeling~\cite{sankaranarayanan2008object,coates2010multi}, and \etc, to aggregate information from multiple views. 

Afterward, when deep learning became popular, people started introducing multi-view object detection methods based on neural networks.
MVDet~\cite{hou2021multiview} introduces an end-to-end method, based on convolutional neural networks, which projects dense 2D image features from cameras to a unified ground plane. After spatial aggregation on the projected features by ground plane convolution, MVDet predicts a pedestrian occupancy map to obtain the final detection results. However, MVDet cannot handle different camera settings, \eg, different numbers of cameras or placements, so that cannot be generalized easily to diverse environments. 
BEVFormer~\cite{li2022bevformer,yang2023bevformer} proposes the spatio-temporal transformer to fuse multi-view features into BEV features, which achieves impressive detection accuracy for autonomous driving related applications. The proposed spatio-temporal transformer is based on deformable attention~\cite{zhu2020deformable} and to sample BEV features from 2D image features by grid-based reference points in BEV, which has similar setups with our smart city applications. 
Therefore, in this work, we aim to adapt BEVFormer under MTMC environments, whose cameras are static but with more distributed placements and variations.

\vspace*{-1em}
\paragraph{Multi-Target Multi-Camera Tracking}
Recent MTMC tracking methods can be divided into three categories as shown in~\cref{fig:mtmc_comp}. 
Late multi-view aggregation methods~\cite{hu2006principal,eshel2008homography,bredereck2012data,hsu2020traffic,hsu2021multi} for MTMC tracking usually adopt a two-stage pipeline, \ie, 2D detection and tracklet generation within each camera view followed by tracklet association across all the cameras. Here, some cross-view association methods~\cite{hu2006principal,eshel2008homography} consider certain geometric constraints, \eg, epipolar geometry constraints or multi-view triangulation. In contrast, others~\cite{hsu2020traffic,hsu2021multi} consider both spatial and temporal constraints, \eg, camera link model. Follow-up works demonstrate that spatial association is much easier and more accurate to be done with ReID features in a unified space by geometrical projections~\cite{xu2016multi,xu2017cross,nguyen2022lmgp,cheng2023rest}. LMGP~\cite{nguyen2022lmgp} formulates a spatial-temporal tracking graph, whose nodes are tracklets from the single-camera tracker, and edges include both temporal and spatial distances.
Whereas ReST~\cite{cheng2023rest} first conducts spatial association and then the temporal association across frames using GNNs.

Early multi-view aggregation for MTMC tracking is first proposed by EarlyBird~\cite{teepe2024earlybird}. It first projects 2D image features into BEV by calibrated projection matrices, followed by stacking and aggregating into BEV features. A CenterNet-based~\cite{zhou2019objects} decoder is appended to obtain detections and ReID features in BEV. As for tracking, EarlyBird follows the idea proposed by FairMOT~\cite{zhang2021fairmot}, which associates detections in the temporal domain with ReID features for appearance and a Kalman filter~\cite{kalman1960new} for motion prediction. 
However, the detection network in EarlyBird is designed for a fixed multi-camera scene, specifically training and inference are under the same scene and the same camera settings. Therefore, it is not flexible for different environments, different numbers of cameras, or different camera placements, which affects its generalizability. As we will show in~\cref{sec:exp}, EarlyBird is also not reliable for long video clips due to its limited long-term association capabilities.

Therefore, we will address the aforementioned limitations and introduce an early multi-view aggregation based MTMC tracker, which is 1) \textit{generalizable} for different scenes and camera settings; 2) more reliable with \textit{better long-term association} to deal with very long video clips.

\section{MCBLT Framework}
\label{sec:method}

\begin{figure*}
  \centering
    \includegraphics[width=0.9\textwidth]{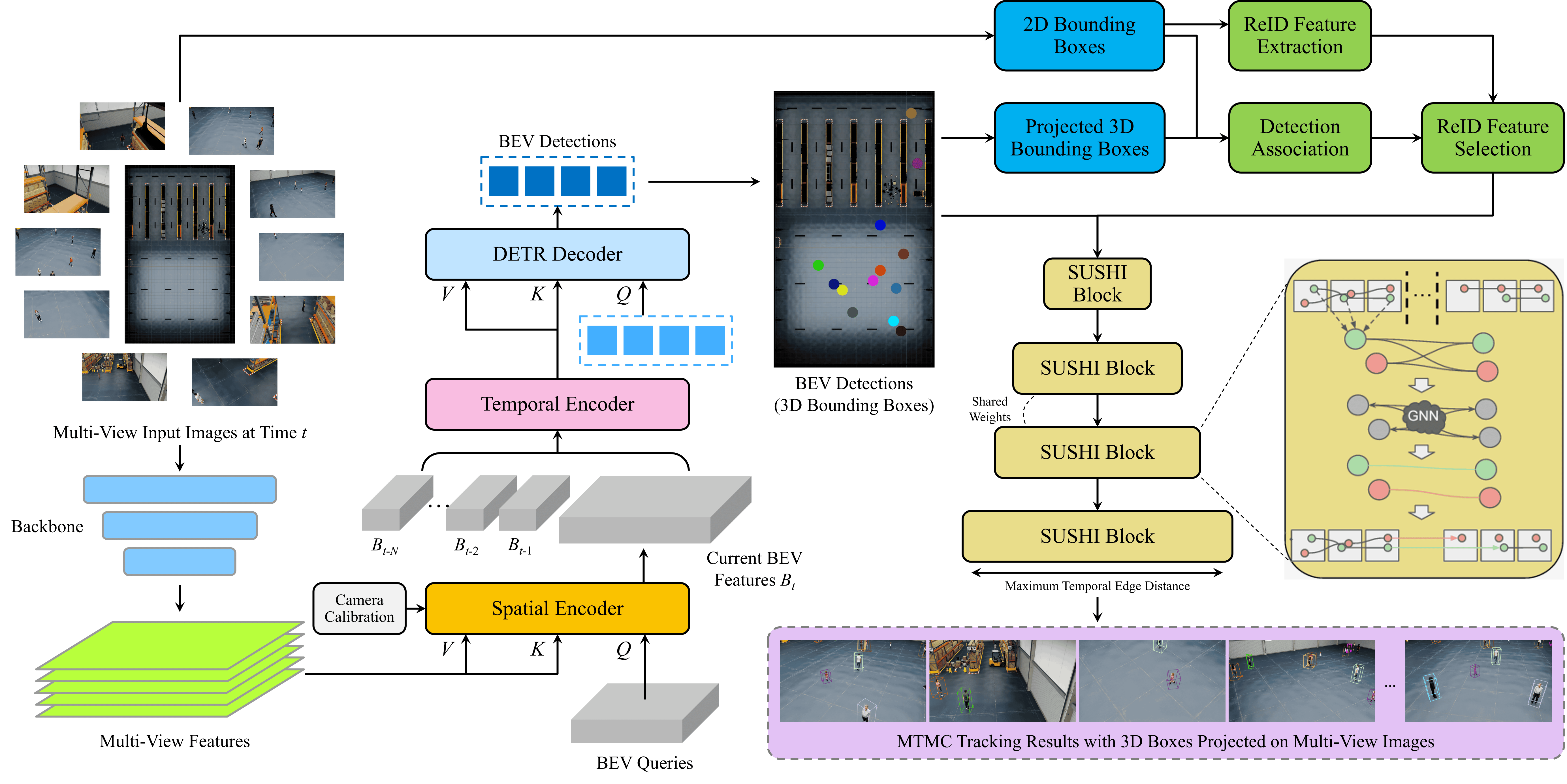}
  \caption{The overall framework of MCBLT. First, multi-view images at frame $t$ are passed through the image backbone to obtain multi-view image features. A spatial encoder is then introduced to aggregate multi-view image features to BEV features $B_t$, followed by a temporal encoder to aggregate BEV features within a temporal window. A DETR-based decoder is utilized to obtain object detection results, which are in the format of 3D bounding boxes. To get reliable ReID features for the detected objects, a ReID feature extraction module is proposed, including a 2D ReID feature extractor and a 2D-3D detection association algorithm. Finally, SUSHI-3D is designed to achieve multi-object tracking in BEV to obtain the final MTMC tracking results. (\textit{SUSHI Block} graphics are from~\cite{cetintas2023unifying}.) }
  \vspace*{-0.5em}
  \label{fig:overall}
\end{figure*}

In this section, we introduce our proposed multi-camera 3D object detection and tracking method in three folds, i.e., 3D object detection, ReID feature extraction, and multi-object tracking in BEV. The overall pipeline is shown in~\cref{fig:overall}.

\subsection{Coordinate Systems and Projection}
\label{subsec:coords}

We first define the 3D world coordinates $\mathcal{W}$ and 3D camera coordinates $\left\{ \mathcal{C}^i \right\}_{i=1}^{V}$ for each specific scene. Besides, we also define the 2D image pixel coordinates $\left\{ \mathcal{U}^i \right\}_{i=1}^{V}$ for each camera image. Here, $V$ is the number of camera views in the scene. The goal is to project multi-view camera information to the 3D world and perform detection and tracking in $\mathcal{W}$ without further cross-view spatial association. 

The origin of $\mathcal{W}$ is defined around the center of the scene, lying on the ground plane. Axes $x$ and $y$ are parallel to the ground plane, and $z$ is vertically pointing up. 
The origin of $\mathcal{C}^i$ is defined at the camera center. Axis $x$ points right, $y$ points down, and $z$ points forward.
A 3D point $\mathbf{x} = [x, y, z]^{\top}$ in the world coordinates $\mathcal{W}$ can then be projected to the pixel coordinates $\mathcal{U}^i$ in the $i$-th camera view as $\mathbf{u} = [u, v]^{\top}$. This projection can be done by
\begin{equation}
    s
    \begin{pmatrix}
        \mathbf{u}\\
        1
    \end{pmatrix}
    =
    \mathbf{P}^i
    \begin{pmatrix}
        \mathbf{x}\\
        1
    \end{pmatrix}
    =
    \mathbf{K}^i \left[ \mathbf{R}^i | \mathbf{t}^i \right]_{3 \times 4}
    \begin{pmatrix}
        \mathbf{x}\\
        1
    \end{pmatrix},
    \label{eq:proj}
\end{equation}
where $s$ is the scale factor, $\mathbf{P}^i$ is the projection matrix, $\mathbf{K}^i$ is the intrinsic matrix, and $\mathbf{R}^i$, $\mathbf{t}^i$ are rotation and translation.

\subsection{Multi-View 3D Object Detection}
\label{subsec:bevformer}

To aggregate multi-view image information in the early stage, we consider projecting image features into bird's-eye view (BEV) and conducting object detection in 3D space. We leverage BEVFormer~\cite{li2022bevformer}, a 3D object detector from surrounding camera views developed for autonomous driving applications, with adaption in~\cref{subsec:datasets}. The key idea of aggregating multi-view information in BEVFormer is to sample $N_{\text{ref}}$ 3D reference points in BEV coordinates, and project these reference points to 2D camera views to sample features from the corresponding image feature maps.

Towards this end, multi-view images $\left\{ I^{i}_t \right\}_{i=1}^V$ at frame $t$ are input into the image backbone to obtain multi-view image features $\left\{ F^{i}_t \right\}_{i=1}^V$.  
The spatial encoder then aggregates $\left\{ F^{i}_t \right\}_{i=1}^V$ to BEV features $B_t$. We adopt the spatial cross-attention (SCA)~\cite{li2022bevformer} for this aggregation,
\begin{equation}
    \text{SCA}(Q_p, F_t) = 
    \frac{1}{|\mathcal{V}_{\text{hit}}|} 
    \sum_{i \in \mathcal{V}_{\text{hit}}} 
    \sum_{j=1}^{N_{\text{ref}}} 
    \text{Attn}(Q_p, \mathcal{P}(p,i,j), F^i_t).
    \label{eq:sca}
\end{equation}
Here, $Q \in \mathbb{R}^{H \times W \times C}$ represents BEV queries, which are pre-defined learnable parameters used as queries. $Q_p \in \mathbb{R}^{C}$ is the query at $p = (x, y)$ in the BEV plane. $\mathcal{V}_{\text{hit}}$ represents the camera views, where the projected 2D reference points fall in. $i$ is the index of the camera view, and $j$ is the index of the reference points. $\text{Attn}(\cdot)$ is the deformable attention layer proposed in~\cite{zhu2020deformable}. $\mathcal{P}(p,i,j)$ is to project the $j$-th 3D reference point to the $i$-th camera view. The 3D to 2D projection is described in~\cref{eq:proj}.

Afterward, a temporal BEV encoder~\cite{yang2023bevformer} is adopted to better incorporate temporal information. Here, a temporal self-attention (TSA) layer~\cite{li2022bevformer} is utilized to gather the BEV feature history,
\begin{equation}
    \text{TSA}(Q_p, {Q, B_{t-1}}) = 
    \sum_{\mathcal{S} \in {Q, B_{t-1}}} 
    \text{Attn}(Q_p, p, \mathcal{S}),
    \label{eq:tsa}
\end{equation}
where $Q_p$ represents the BEV query located at $p=(x, y)$.

Finally, a deformable DETR head~\cite{zhu2020deformable} is followed to predict 3D bounding boxes from BEV features provided by the temporal encoder. We use Focal loss for classification training and $L_1$ loss for bounding box regression supervision.

\subsection{Multi-View ReID Feature Extraction}
\label{subsec:reid}

\begin{figure}[t]
  \centering
   \includegraphics[width=0.9\linewidth]{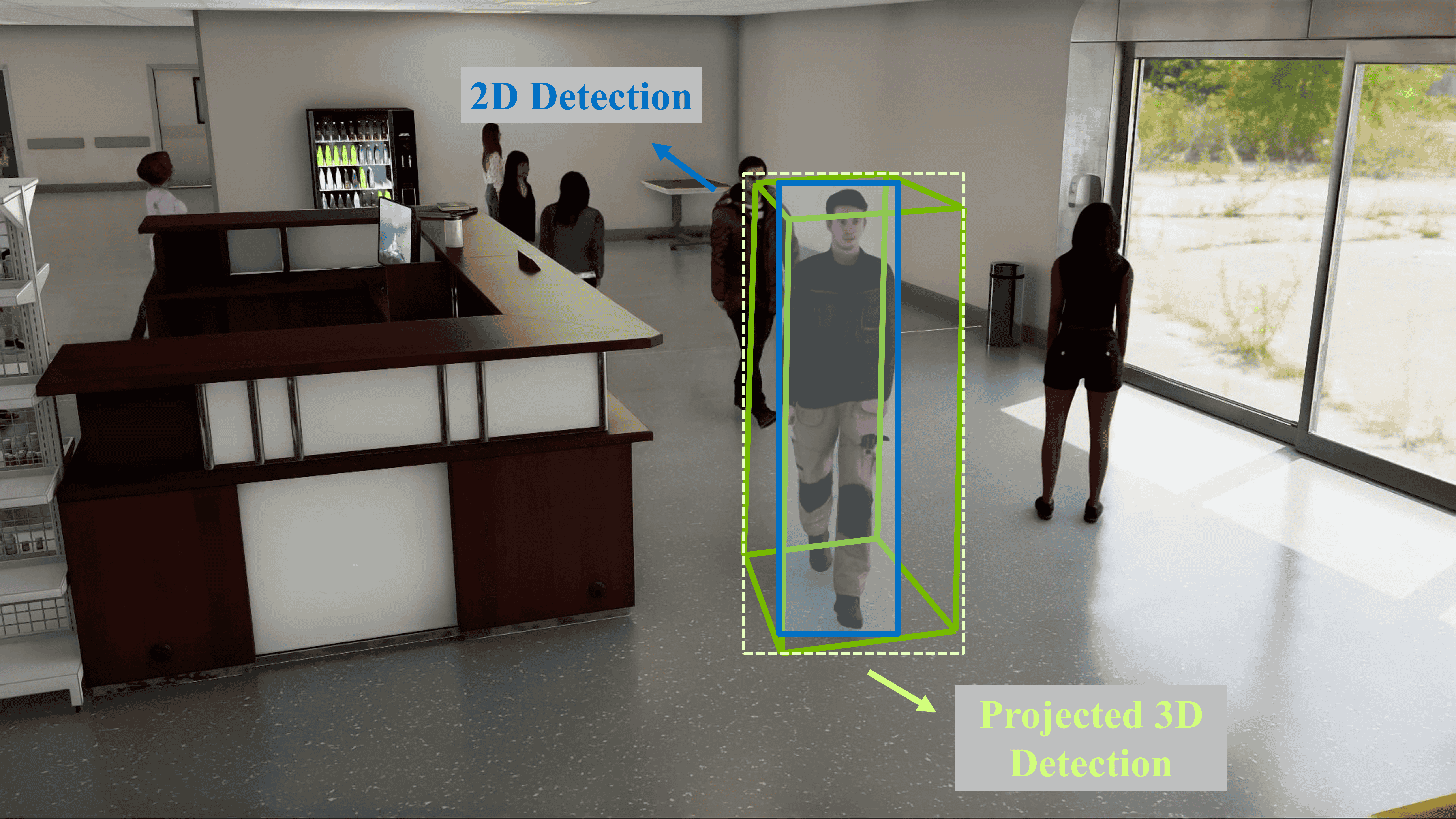}
   \caption{Illustration on 2D detection (in blue) and the corresponding projected 3D detection (in green).}
   \vspace*{-1em}
   \label{fig:det_assoc_example}
\end{figure}

Appearance features are crucial for tracking tasks, especially for cross-view tracking, since object appearances from different cameras might differ due to varied illuminations and viewing angles. The na\"ive way to obtain the appearance feature for each 3D detection from BEVFormer is to project the 3D bounding box to each camera image and extract ReID features by the projected box.

However, the projected 3D bounding boxes are usually much larger than the actual objects in the image (see~\cref{fig:det_assoc_example}). 
This will include noisy background or other nearby objects in the target object so that affect the quality of the ReID features.
Therefore, we propose a 2D-3D detection association algorithm to effectively find the best 2D bounding box for each 3D detection.

With 3D detections and ReID features of each 2D detection, we then assign each 3D bounding box with one or more ReID features, by introducing a 2D-3D detection association algorithm to create a mapping from the 3D bounding box set to the 2D bounding box set. 
First, we project 3D bounding boxes to 2D in each camera view, then we compute a cost matrix between 2D detections and projected 3D detections by
\begin{equation}
    c_{ij} = 
    \left\{
    \begin{aligned}
    & \lambda \left\lVert \mathbf{bc}_{i}^{\text{3D}} - \mathbf{bc}_{j}^{\text{2D}} \right\rVert_2, 
    & & \text{if } IOU(\mathbf{b}_{i}^{\text{3D}}, \mathbf{b}_{j}^{\text{2D}}) \geq 0.1, \\
    & +\infty, & & \text{otherwise},
    \end{aligned}
    \right.
    \label{eq:assoc_cost}
\end{equation}
where $\mathbf{bc}^{\text{2D}}$ is the bottom center point of the 2D bounding box, $\mathbf{bc}^{\text{3D}}$ is the bottom center point of the 2D area of the projected 3D bounding box, and $\lambda$ is a robustness factor to handle 2D occlusions 
\begin{equation}
    \lambda = 
    \mathbbm{1}(v_{i}^{\text{3D}} \geq v_{j}^{\text{2D}}) + 
    \alpha \mathbbm{1}(v_{i}^{\text{3D}} < v_{j}^{\text{2D}}).
\end{equation}
Here, we set $\alpha > 1$ to penalize if $\mathbf{b}^{\text{2D}}$ is lower than $\mathbf{b}^{\text{3D}}$. 
We utilize the Hungarian algorithm to achieve the final assignment. Moreover, the 3D detections can be verified here through the following strategy: a 3D detection will be removed if it cannot be matched with any 2D detections from all camera views.
The detailed algorithm is described in~\cref{subsec:supp_det_assoc_alg}.

\subsection{3D Multi-Object Tracking with GNNs}
\label{subsec:sushi3d}
The early multi-view aggregation of our MCBLT framework allows us to perform multi-object tracking (MOT) directly in 3D world coordinates $\mathcal{W}$.
Graphs provide a natural framework to address the long-term association challenges in MTMC tracking within a tracking-by-detection setting.
We build on SUSHI~\cite{cetintas2023unifying}, originally designed for 2D tracking, to introduce the first hierarchical GNN-based tracking solution in 3D space. To this end, we first extend its graph formulation and features to 3D. Then we tackle the weaknesses of SUSHI’s long-term tracking, i.e., \textit{heuristic matching} of overlapping windows and occlusion handling being limited to the \textit{window size} by introducing a novel global block to the hierarchy. Overall, we demonstrate long-term tracking on a previously unseen temporal scale (SUSHI: 512 frames vs. MCBLT full sequence).

\vspace*{-1em}
\paragraph{MOT graph formulation.}
In the common graph formulation of multi-object tracking~\cite{network_flows_tracking}, each node $v \in V$ models a detection connected with edges $e \in E$ representing association hypotheses in an undirected graph $G=(V, E)$.
Learning or optimizing the graph connectivity results in object tracks $\mathcal{T}$ across time.
Nodes and edges are initialized with object identity-relevant information projected to the embedding space.
These embeddings encode geometrical, appearance, and/or motion features to infer the object identity via graph connection.
More specifically, edges contain distance information between their connecting nodes, \eg, cosine distances of ReID features.
With the help of message passing, first introduced to MOT by~\cite{braso_2020_CVPR}, information is shared and distributed across the graph.
The procedure iteratively updates the node $h_i$ and edge $h_{(i, j)}$ embeddings via the following rules
\begin{align}
&(v\rightarrow e) &h_{(i, j)}^{(l)} &= \mathcal{N}_e\left(\left[h_i^{(l-1)}, h_j^{(l-1)}, h_{(i, j)}^{(l-1)}\right]\right),  \label{node2edge}\\
&(e\rightarrow v) 
            &m_{(i, j)}^{(l)} &=  \mathcal{N}_v\left(\left[h_i^{(l-1)}, h_{(i, j)}^{(l)} \right]\right),  \label{edge2node1} \\
&            &h_i^{(l)} &= \Phi \left(\left \{ m_{(i, j)}^{(l)}   \right\}_{ j \in N_i}\right),  \label{edge2node2} 
\end{align}
with $\mathcal{N}_e$ and $\mathcal{N}_v$ representing learnable functions, $[\cdot]$ denotes concatenation, $N_i\subset V$ is the set of adjacent nodes to $i$, and $\Phi$ denotes summation, maximum or an average.
Finally, a binary edge classification and linear program that ensures the network flow integrity (single edge in and out of a node) yields identity-consistent object trajectories. 

To model long-term object associations across thousands of frames in a computationally feasible manner, we follow prior work to sparsify the fully connected MOT graph via an initial graph pruning~\cite{braso_2020_CVPR} and a hierarchical model architecture~\cite{cetintas2023unifying}.
During graph construction, the pruning utilizes the same aforementioned object identity cues to remove unlikely edges, \ie, object associations.
This approach not only increases the feasible number of input frames for the graph model but improves overall performance by reducing the imbalance between positive and negative edge classifications.
Splitting the input frames into non-overlapping sub-graphs processed by a hierarchy of GNNs boosts the modeling capacities even further.
To this end, the graph formulation is extended to nodes representing tracklets and each hierarchy level merges nodes from previous levels, thereby, providing the inputs for subsequent levels.

\vspace*{-1em}
\paragraph{Tracking with GNNs in 3D space.}
The generalizable and versatile graph structure as well as node and edge feature design in~\cite{cetintas2023unifying} facilitates its application to 3D space.
For MCBLT, we replace the 2D geometry, appearance, and motion feature encodings with MTMC 3D counterparts.

For the \emph{geometry} embeddings, we encode $xyz$-center distances for an edge connecting nodes $i$ and $j$ via
\begin{equation}
    \left( x_i - x_j, y_i - y_j, z_i - z_j  \right).
\end{equation}

In contrast to 2D bounding boxes, geometric distances in 3D are not impacted by projection distortions and camera distance scaling.

MCBLT extracts \emph{appearance} information in the form of multi-view ReID features obtained via 3D-2D projection and detection association (described in~\cref{subsec:reid}).
To provide our graph model with view-consistent and stable appearance cues, we compute cosine distances of ReID features averaged across all cameras in which the object is observable.

In contrast to 2D tracking, our strong combination of true-to-scale 3D geometry and multi-view appearance renders the impact of linear \emph{motion} features negligible.
Hence, we opted for a more efficient graph formulation without motion features but with larger input time spans.
We leave the exploration of more sophisticated motion models for 3D GNN tracking open for future research.

\vspace*{-1em}

\begin{figure}
  \centering
    \includegraphics[width=0.9\linewidth]{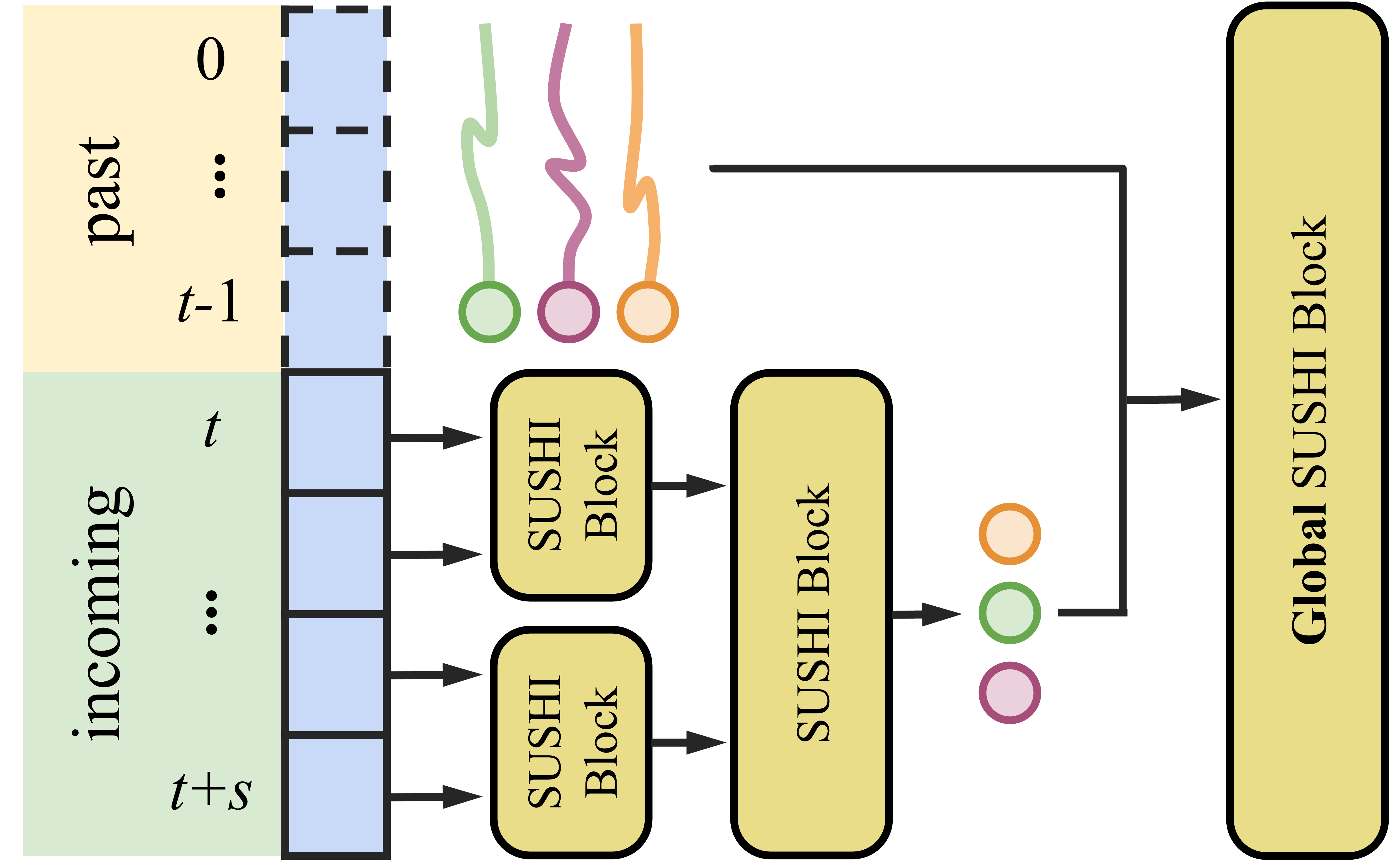}
  \caption{GNN hierarchy of MCBLT for tracking. We process long sequences in a near-online fashion with stride $s$. But in contrast to~\cite{cetintas2023unifying}, we omit overlaps between windows and heuristics. To associate incoming with past tracks, MCBLT uses a global merging block. The global block requires no additional training. }
  \vspace*{-1em}
  \label{fig:sushi_near_online_inference}
\end{figure}

\paragraph{Long-term tracking with global block.}
Due to their larger camera coverage, multi-camera setups can observe objects across longer time spans and, therefore, pose uniquely challenging long-term tracking problems. 
For example, AICity'24~\cite{wang20248th} consists of sequences with up to 24k frames including object occlusions for up to 2k frames, a magnitude longer compared to popular single-view tracking benchmarks~\cite{motchallenge}
The original SUSHI~\cite{cetintas2023unifying} inference computes overlapping graph outputs and associates tracks via handcrafted and, therefore, error-prone matching heuristics.

In this work, we propose a novel near-online inference and global tracking block that associates previous tracks with predictions on the incoming frames.
Figure~\cref{fig:sushi_near_online_inference} depicts our new inference without graph overlaps.
Our GNN hierarchy first processes the non-overlapping incoming frames with regular SUSHI blocks and then associates objects across the full sequence with the final global tracking block.
We process the entire sequence in a sliding window fashion.
%

%
The global block shares weights with the previous hierarchy level and thus requires no additional training.
Our MCBLT removes potential biases in previous heuristics in GNN-based tracking and unlocks occlusion handling beyond the number of frames per graph.

\section{Experiments}
\label{sec:exp}

\subsection{Datasets}
\label{subsec:datasets}

\paragraph{AICity'24 Dataset~\cite{wang20248th}} 
is an MTMC tracking benchmark consisting of 6 different synthetic environments, \eg, warehouses, retail stores, and hospitals, developed using the NVIDIA Omniverse Platform~\cite{nvidia_omniverse}. The dataset includes 90 scenes, 40 for training, 20 for validation, and 30 for testing, with a total of 953 cameras, 2,491 people, and over 100 million bounding boxes. 
Besides, we also include 30 additional scenes with similar camera settings for BEVFormer training. In this dataset, persons are annotated as 3D bounding boxes with 3D locations, dimensions, and orientations. Long-term tracking performance is crucial for AICity'24~\cite{wang20248th} as it evaluates identity consistency across sequences with up to 24k frames including occlusions ranging up to 2k frames.

\vspace*{-1em}
\paragraph{WildTrack Dataset~\cite{chavdarova2018wildtrack}}
is a real-world MTMC tracking benchmark containing a single sequence of a scene where 7 synchronized cameras cover over a 36$\times$12 m space at a 1920$\times$1080 resolution. There are 400 fully annotated frames per camera at 2 FPS with a total of 313 identities and 42,721 2D bounding boxes. As for the 3D annotations, there is no 3D bounding box available. A grid is defined on the ground plane and all persons are annotated by the corresponding grid indices, which can be mapped to 3D locations on the ground plane in meters. Following the experiment setting in \cite{hou2021multiview,cheng2023rest,teepe2024earlybird}, we use the first 360 annotated frames as the training data and evaluate the rest 40 frames. 

\subsection{Implementation Details}
\paragraph{BEVFormer adaption for MTMC environments.}
The original BEVFormer is developed for autonomous driving applications with moving cameras but fixed camera settings, \ie, a fixed number of cameras and fixed relative camera placement. For MTMC environments, we usually have dynamic camera systems across different scenes. AICity'24 dataset, for example, has 6 different indoor environments with cameras ranging from 7 to 16. Therefore, we re-design the spatial cross-attention layers to enable a dynamic number of cameras for different scenes and remove camera embeddings. Besides, we apply a transformation on the original 3D world coordinates to shift the origin to the center of the scene (see \cref{subsec:supp_exp_det3d_recenter}). Moreover, we implement an additional Circle Non-Maximum Suppression (NMS) with a threshold of 0.2m to filter nearby false positives. 
Since there is no 3D bounding box annotation in the WildTrack dataset, we remove the bounding box dimension and orientation regression losses during BEVFormer training. We use a default 3D bounding box dimension (width, length, height) of $[0.6, 0.6, 1.7]$ and zero rotation for later 2D-3D detection association.

\vspace*{-1em}
\paragraph{2D detector.}
We use DINO~\cite{zhang2022dino} with FAN-small backbone~\cite{zhou2022understanding}. The detector is pre-trained on a subset (800k images) of the OpenImages dataset~\cite{kuznetsova2020open}, pseudo-labeled by a DINO detector trained on 80 COCO classes. Then, the detector is fine-tuned on a proprietary dataset with more than 1.5 million images and more than 27 million objects for the person class.

\vspace*{-1em}
\paragraph{ReID feature extraction.}
We extract ReID features for each 2D bounding box from the 2D detector. Our person-centric ReID model is based on the SOLIDER~\cite{chen2023beyond}, a self-supervised learning framework with a Swin-Tiny Transformer~\cite{liu2021Swin} backbone. The model uses image crops of size 256$\times$128 as input and outputs a feature of dimension 256 enhancing memory efficiency and throughput. We pre-train the model on a proprietary dataset of 3 million unlabeled image crops. The supervised fine-tuning is performed on a collection of both real and synthetic datasets which includes characters from Market-1501~\cite{zheng2015scalable}, AICity'24~\cite{wang20248th} and additional propitiatory datasets totaling the object count to 3,826 identities and image count to 82k object crops. We analyze the ReID feature quality in~\cref{sec:supp_reid}.

\subsection{Results on AICity'24 Dataset}
\label{subsec:res_aic24}

\begin{figure*}
  \centering
    \includegraphics[width=0.95\textwidth]{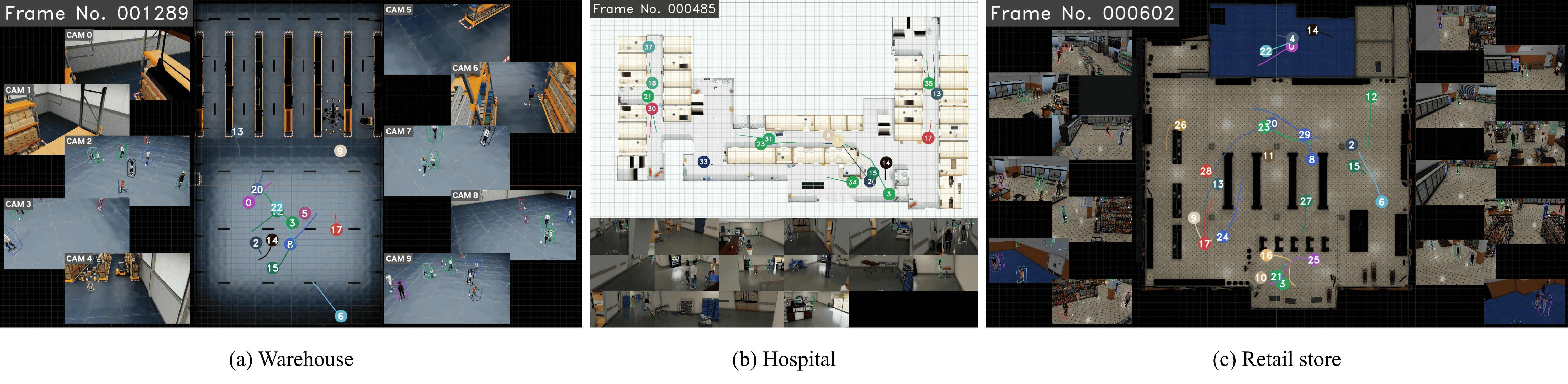}
    \vspace*{-0.4cm}
  \caption{Visualization of MTMC detection and tracking results for three different scenes in AICity'24 test set. The tracked objects are shown as colored dots in the BEV floor plans, and object 3D bounding boxes are projected and drawn in each camera view.}
  \vspace*{-0.5em}
  \label{fig:viz_aic24}
\end{figure*}

The AICity'24 benchmark adopts the Higher Order Tracking Accuracy (HOTA) scores~\cite{luiten2020IJCV} for evaluation. HOTA is computed on the 3D locations of objects, with repetitive data points removed across cameras for the same frame. Euclidean distances between predicted and ground truth 3D locations are converted to similarity scores using a zero-distance parameter. These scores contribute to the calculation of localization accuracy (LocA), detection accuracy (DetA), and association accuracy (AssA). 

\begin{table}
  \centering
  \small
  \begin{tabular}{@{\hspace{1mm}}l|cccc@{\hspace{1mm}}}
    \toprule
    Method & HOTA$\uparrow$ & DetA$\uparrow$ & AssA$\uparrow$ & LocA$\uparrow$ \\
    \midrule
    Asilla~\cite{AICity24Paper43} & 40.34 & 53.80 & 32.50	& 89.57 \\
    ARV~\cite{AICity24Paper40} & 51.06 & 54.85 & 48.07 & 89.61 \\
    UW-ETRI~\cite{AICity24Paper6} & 57.14 & 59.88 & 54.80 & 91.24 \\
    FraunhoferIOSB~\cite{AICity24Paper46} & 60.88 & 69.54 & 55.20 & 87.97 \\
    Nota~\cite{AICity24Paper39} & 60.93 & 68.37 & 54.96 & 90.62 \\
    SJTU-Lenovo~\cite{AICity24Paper1} & 67.22 & \underline{84.03} & 55.06 & \underline{93.82} \\
    Yachiyo~\cite{AICity24Paper32} & \underline{71.94} & 72.10 & \underline{71.81} & 88.39 \\
    \midrule
    MCBLT (Ours) & {\bf 81.22} & {\bf 86.94} & {\bf 76.19} & {\bf 95.67} \\
    \bottomrule
  \end{tabular}
  \caption{Results on AICity'24 test set. The first place is in \textbf{bold}, and the second place is \underline{underlined}.}
  \label{tab:res_aic24}
\end{table}

MCBLT reaches SOTA on the AICity'24 benchmark on all metrics with improvements of $+9.28$ on HOTA, $+2.91$ on DetA, $+4.38$ on AssA, and $+1.85$ on LocA. These significant improvements demonstrate MCBLT's effectiveness in advancing multi-camera tracking performance.
We present qualitative results on the AICity'24 dataset in~\cref{fig:viz_aic24} by projecting 3D detections with track IDs to each camera view and plot their 3D locations on the floor plans.
Overall, our quantitative and qualitative results show that MCBLT achieves robust MTMC detection and tracking accuracies in various indoor environments.

Besides, we evaluate the 2D-3D detection association algorithm to ensure we have high-quality ReID features assigned to each 3D detection. This evaluation is based on the association results between ground truth 2D and 3D annotations in the AICity'24 training set. We introduce the evaluation metric, detection association accuracy, defined as the number of corrected matches divided by the number of matched ground truth 2D bounding boxes. The evaluation results are shown in~\cref{tab:det_assoc_eval}. We can find that even in crowded hospital scenes, our algorithm can achieve above 90\% accuracy, illustrating its robustness. More visualizations based on predictions are shown in~\cref{subsec:supp_det_assoc_viz}.

\begin{table}
  \centering
  \small
  \begin{tabular}{l|ccc|c}
    \toprule
    Scene & Warehouse & Retail store & Hospital & Overall \\
    \midrule
    Accuracy & 99.4\% & 95.2\% & 91.2\% & 96.9\% \\
    \bottomrule
  \end{tabular}
  \caption{Detection association accuracy among different scene types on AICity'24 dataset.}
  \vspace*{-1em}
  \label{tab:det_assoc_eval}
\end{table}

\subsection{Results on WildTrack Dataset}
\label{subsec:res_wildtrack}

The WildTrack dataset evaluates tracking results in the ground plane with a 1m threshold for GT assignment. The primary metrics are IDF1~\cite{ristani2016performance}, Multi-Object Tracking Accuracy (MOTA), and Mutli-Object Tracking Precision (MOTP)~\cite{bernardin2008evaluating}. Furthermore, we report the number of Mostly Tracked (MT) and Lost (ML) tracks in percentages.

\begin{figure*}
    \includegraphics[width=0.95\textwidth]{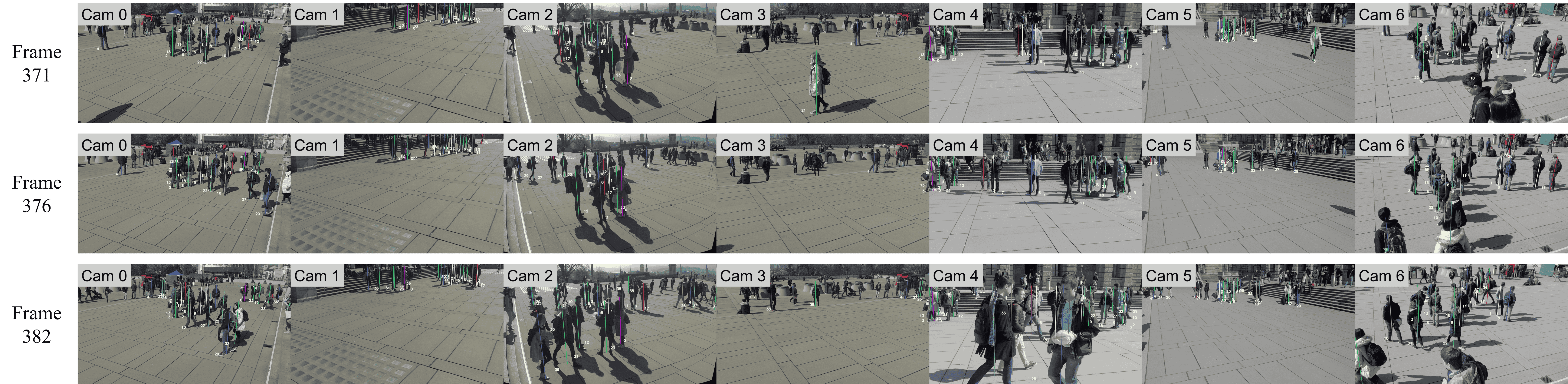}
  \caption{Visualization of MTMC detection and tracking results on WildTrack test set. The tracked objects are shown as colored poles projected in each camera view. The poles are defined by their 3D foot points and a pre-defined human height (1.7m).}
  \vspace*{-1em}
  \label{fig:viz_wildtrack}
\end{figure*}

\begin{table}
  \centering
  \small
  \begin{tabular}{@{\hspace{1mm}}l|C{0.6cm}C{0.7cm}C{0.7cm}C{0.5cm}C{0.5cm}@{\hspace{2mm}}}
    \toprule
    Method & IDF1$\uparrow$ & MOTA$\uparrow$ & MOTP$\uparrow$ & MT$\uparrow$ & ML$\downarrow$ \\
    \midrule
    KSP-DO~\cite{chavdarova2018wildtrack} & 73.2 & 69.6 & 61.5 & 28.7 & 25.1 \\
    KSP-DO-ptrack~\cite{chavdarova2018wildtrack} & 78.4 & 72.2 & 60.3 & 42.1 & 14.6 \\
    GLMB-YOLOv3~\cite{ong2020bayesian} & 74.3 & 69.7 & 73.2 & 79.5 & 21.6 \\
    GLMB-DO~\cite{ong2020bayesian} & 72.5 & 70.1 & 63.1 & {\bf 93.6} & 22.8 \\
    DMCT~\cite{you2020real} & 77.8 & 72.8 & 79.1 & 61.0 & \underline{4.9} \\
    DMCT Stack~\cite{you2020real} & 81.9 & 74.6 & 78.9 & 65.9 & \underline{4.9} \\
    ReST~\cite{cheng2023rest} & 86.7 & 84.9 & 84.1 & 87.8 & \underline{4.9} \\
    EarlyBird~\cite{teepe2024earlybird} & 92.3 & \underline{89.5} & 86.6 & 78.0 & \underline{4.9} \\
    \midrule
    MCBLT (Ours) & \underline{93.4} & 87.5 & {\bf 94.3} & \underline{90.2} & {\bf 2.4} \\
    MCBLT$^{\dagger}$ (Ours) & {\bf 95.6} & {\bf 92.6} & \underline{93.7} & 80.5 & 7.3 \\
    \bottomrule
  \end{tabular}
  \caption{Results on WildTrack test set. The first place is in \textbf{bold}, and the second place is \underline{underlined}. $\dagger$ uses the same detections as EarlyBird~\cite{teepe2024earlybird}.}
  \vspace*{-1em}
  \label{tab:res_wildtrack}
\end{table}

WildTrack only provides limited training data (360 frames), while transformer-based networks, \eg, BEVFormer, generally require larger amounts of data to unfold their potential. Thus, to have a fair comparison, we report our results under two settings: (i) MCBLT with detections from BEVFormer, (ii) MCBLT$^{\dagger}$ using the same detections as EarlyBird (based on MVDet~\cite{hou2021multiview} with ResNet-18 backbone). MCBLT achieves the state-of-the-art with $+1.1$ IDF1 improvement over EarlyBird~\cite{teepe2024earlybird} even if it reaches a slightly lower MOTA score (as expected due to the transformer backbone). MCBLT$^{\dagger}$ significantly outperforms EarlyBird with $+3.3$ IDF1 and $+3.1$ MOTA while using the same detections, demonstrating the efficacy of our methods.

We provide qualitative results on WildTrack in~\cref{fig:viz_wildtrack}. In addition, we present additional results on WildTrack by pre-training BEVFormer on the large-scale synthetic dataset (AICity'24) and finetuning on the small WildTrack dataset in the supplementary material~\cref{subsec:supp_exp_det3d_pretrain}.

\subsection{Ablation Studies}
\label{subsec:ablation}
\paragraph{Multi-view object detector configuration analysis.} To verify we have the best BEVFormer model for multi-view 3D object detection, we conduct several ablation studies on backbone selection and parameter tuning to improve the detection results. The results are shown in~\cref{tab:det_config}.
We use a customized validation set from the AICity'24 dataset, including around 3k frames from warehouse, hospital, and retail store scenes. More results shown in~\cref{subsec:supp_exp_det3d_detectors}.
According to the results in~\cref{tab:det_config}, we use ResNet-101 with higher BEV resolution as our 3D detector.

\begin{table}
  \centering
  \small
  \begin{tabular}{lcccc}
    \toprule
    Backbone & BEV Reso. & Voxel Size & Epochs & mAP \\
    \midrule
    ResNet-50 & 50$\times$50 & 2.0 m & 24 & 83.14 \\
    ResNet-101 & 200$\times$200 & 0.5 m & 24 & 88.64 \\
    ResNet-101 & 200$\times$200 & 0.5 m & 48 & 95.36 \\
    \bottomrule
  \end{tabular}
  \caption{Ablation studies on the configurations of our multi-view object detector. The evaluation is done on the customized validation set of the AICity'24 dataset.}
  \vspace*{-1em}
  \label{tab:det_config}
\end{table}

\vspace*{-1em}
\paragraph{Long-term association analysis.}
The length of sequences and occlusion gaps common in MTMC benchmarks without full camera coverage of the environment pose unique challenges to our tracking model and inference.
To process arbitrarily long sequences, SUSHI~\cite{cetintas2023unifying} uses heuristics to stitch overlapping graphs in a sliding window fashion. 
In~\cref{tab:sushi_eval_ablation}, we compare the heuristic matching of SUSHI with our global block for long-term association. 
Rows 1-4 demonstrate how increasing the window size gradually improves the association performance, \ie, HOTA and AssA.
The last row replaces all heuristics with our novel global merging block while using the \textit{same learnable weights} as row 4. 
Our global merging block increases overall performance by $+4.42$ HOTA by scaling the association window to the full length of the video.

To further illustrate the superiority of our learned GNN approach, we evaluate progressively longer subsets of a selected AICity'24 scene in~\cref{tab:long_term}.
The BEV-KF baseline represents heuristic online trackers like EarlyBird~\cite{teepe2024earlybird} commonly only evaluated on short tracking challenges like WildTrack.
From 1,000 frames to the full AICity'24 length, MCBLT only drops by $4.58$ HOTA points compared to the massive performance decrease of $43.35$ utilizing only a Kalman filter.
Such trackers fail to tackle the long-term tracking challenges common in MTMC benchmarks.

\begin{table}
  \centering
  \small
  \begin{tabular}{c | c | ccc}
    \toprule
    Long-term Asc. & Max. Window & HOTA & AssA & DetA \\
    \midrule
    \multirow{4}{*}{Heuristics~\cite{cetintas2023unifying}} & 480 & 51.87 & 31.26 & 86.12 \\
    & 960 & 62.50 & 45.25 & 86.36 \\
    & 1920 & 71.88 & 59.78 & 86.47 \\
    & 3840 & 76.80 & 68.22 & 86.49 \\
    \midrule
    Model (Ours) & Global & 81.22 & 76.19 & 86.94 \\
    \bottomrule
  \end{tabular}
  \caption{
    SUSHI inference ablation on the AICity'24 test set.
    Rows 1 to 4 rely on heuristic matching to track overlapping sliding windows inferred by the SUSHI GNN hierarchy.
    Our final tracking solution (the last row) applies a global merging block to associate without graph overlaps.
  }
  \vspace*{-0.3em}
  \label{tab:sushi_eval_ablation}
\end{table}

\begin{table}
  \centering
  \small
  \begin{tabular}{p{38pt}|c|ccc}
    \toprule
    Method & \# of frames & HOTA & AssA & DetA \\
    \midrule
    \multirow{5}{*}{MCBLT} & 1,000 & 95.09 & 95.77 & 94.41 \\
    & 3,000 & 93.62 & 92.37 & 94.89 \\
    & 5,000 & 91.85 & 88.82 & 94.98 \\
    & 10,000 & 91.06 & 88.33 & 93.89 \\
    & 23,994 & 90.51 & 88.05 & 93.03 \\
    \midrule
    \multirow{5}{\linewidth}{BEV-KF (baseline)} & 1,000 & 63.49 & 45.29 & 89.01 \\
    & 3,000 & 45.28 & 22.86 & 89.71 \\
    & 5,000 & 35.54 & 14.23 & 88.79 \\
    & 10,000 & 27.91 & 8.87 & 87.80 \\
    & 23,994 & 20.15 & 4.67 & 86.91 \\
    \bottomrule
  \end{tabular}
  \caption{Tracking comparisons with increasing video lengths from a warehouse scene in AICity'24 dataset. The baseline BEV-KF is based on the detection results from BEVFormer processed by a Kalman filter based tracker~\cite{wang2019jde} used in EarlyBird~\cite{teepe2024earlybird}.}
  \vspace*{-1em}
  \label{tab:long_term}
\end{table}

\section{Conclusion}
\label{sec:conclusion}

This paper proposed MCBLT, an accurate and robust multi-camera 3D detection and tracking framework, based on early multi-view aggregation, in environments monitored by static multi-camera systems. The proposed framework has impressive generalizability for diverse scenes and camera settings. It also has a powerful capability for long-term association to track objects in very long videos.
As a result, MCBLT achieves SOTA on both the AICity'24 dataset and the WildTrack dataset.

{
    \small
    \bibliographystyle{ieeenat_fullname}
    \bibliography{main}
}

\appendix
\clearpage
\maketitlesupplementary

\section{Experiments on Mutli-View 3D Detector}
\label{sec:supp_exp_det3d}

\subsection{3D Object Detector Comparisons}
\label{subsec:supp_exp_det3d_detectors}

We conducted experiments on different settings for our 3D object detector in~\cref{tab:supp_det3d}. We considered both BEVFormer~\cite{li2022bevformer} and BEVFormer v2~\cite{yang2023bevformer} with ResNet-50, ResNet-101, and V2-99 backbones. The maximum number of cameras for training is reported for each experiment due to the limitation of H100 GPU memory. 
All experiments were trained for 24 epochs with a learning rate of $2\times10^{-4}$.

\begin{table}[h]
  \centering
  \small
  \begin{tabular}{@{\hspace{1mm}}llC{1.2cm}|C{1cm}@{\hspace{1mm}}}
    \toprule
    Method & Backbone & Max \# of Cameras & mAP \\
    \midrule
    BEVFormer~\cite{li2022bevformer} & ResNet-50 & 16 & 83.14 \\
    BEVFormer~\cite{li2022bevformer} & ResNet-101 & 15 & 88.64 \\
    BEVFormer v2~\cite{yang2023bevformer} & ResNet-50 & 15 & 82.78 \\
    BEVFormer v2~\cite{yang2023bevformer} & ResNet-101 & 15 & 85.03 \\
    BEVFormer v2~\cite{yang2023bevformer} & V2-99 & 14 & 79.95 \\
    \bottomrule
  \end{tabular}
  \caption{3D object detection results with different detectors on a customized AICity'24 validation set.}
  \vspace*{-1em}
  \label{tab:supp_det3d}
\end{table}

Compared with BEVFormer, BEVFormer v2 receives relatively lower mAP by adding the perspective supervision. Therefore, the perspective supervision may not be helpful for our MTMC application for model convergence.
As for the V2-99 backbone, we need to decrease the number of camera views during the training to fit our GPU memory of around 80 GB. This will downgrade the detection performance significantly. 
In the future, we will improve the memory efficiency to make it possible to utilize larger image backbones.



\subsection{Scene Re-Centering for BEVFormer}
\label{subsec:supp_exp_det3d_recenter}

The definition of the BEV coordinate system is important for BEVFormer training. In the original autonomous driving settings, the origin is located on the ego-vehicle, which is the center of the area to be perceived. In our MTMC settings, we define the origins of the multi-camera scenes as the centers of the floor plans and transform the annotations and calibration matrices to the newly defined BEV coordinates. We call this step ``re-centering''. 

In~\cref{tab:supp_wildtrack}, we evaluate the model performance on the WildTrack dataset before and after this re-centering step. Before the re-centering, the origin was defined at the corner of a scene.
We notice that re-centering can dramatically improve the detection performance by $+22.33$ mAP, especially for those objects farther from the origin.

\subsection{Pre-Training on AICity'24 Dataset}
\label{subsec:supp_exp_det3d_pretrain}

Since WildTrack is a small dataset with only 400 frames in total, we considered training BEVFormer with a pre-trained model on the AICity'24 dataset, which is a much larger dataset with various scenes. As shown in~\cref{tab:supp_wildtrack}, this pre-training leads to $+3.67$ detection performance improvement on the WildTrack test set.
This also illustrates the important role of large and well-annotated synthetic datasets in boosting the performance on limited real data. 

\begin{table}[t]
  \centering
  \small
  \begin{tabular}{l|c}
    \toprule
    Method & mAP \\
    \midrule
    Baseline & 66.03\\
    + re-centering & 88.36\\
    + pre-training & 92.03\\
    \bottomrule
  \end{tabular}
  \caption{A comparison of detection results on the WildTrack dataset with \textit{re-centering} and \textit{pre-training}.}
  \vspace*{-1em}
  \label{tab:supp_wildtrack}
\end{table}

\section{Detection Association Algorithm}
\label{sec:supp_det_assoc}


\subsection{Algorithm Details}
\label{subsec:supp_det_assoc_alg}

\begin{algorithm}[h]
\SetAlgoLined
\SetKwInOut{Input}{Input}
\SetKwInOut{Output}{Output}
\SetKw{KwFrom}{from}
\SetKw{KwAnd}{and}
\Input{2D detection set $\mathcal{D}^v$ from camera $v$; 3D detection set $\mathcal{E}$ from BEVFormer with all camera views; projection matrix $\mathcal{P}^v$ of camera $v$.}
\Output{Mapping of indices from $\mathcal{E}$ to $\mathcal{D}^v$.}

$\mathcal{E} \leftarrow$ filter $\mathcal{E}$ by confidence score\;
$\mathcal{E} \leftarrow$ CircleNMS$(\mathcal{E}, \delta)$\; \tcp*[f]{\small optional, $\delta$: NMS threshold}

$\mathcal{E}^{v} \leftarrow \mathcal{P}^v (\mathcal{E})$\tcp*[r]{\small projected 3D boxes}
\For{camera $v$ \KwTo $V$}{
  Initialize the cost matrix $\mathbf{c}^v = [c_{ij}^v]$ as zeros\;
  \For{$\mathbf{b}_{i}^{\text{3D}}$ \KwFrom $\mathcal{E}^v$}{
    \For{$\mathbf{b}_{j}^{\text{2D}}$ \KwFrom $\mathcal{D}^v$}{
      $c_{ij}^v \leftarrow$ compute cost by~\cref{eq:assoc_cost}\;
    }
  }
  Matches $m^v \leftarrow$ Hungarian$(\mathbf{c}^v, \Delta)$\; \tcp*[f]{\small $\Delta$: cost threshold}
}
\caption{2D-3D detection association}
\label{alg:det_assoc}
\end{algorithm}

The detailed 2D-3D detection association algorithm is shown in~\cref{alg:det_assoc}.
We set the threshold for CircleNMS to $\delta=0.2$m and set the cost threshold to $\Delta = 150$.

\begin{figure*}[t]
  \centering
    \includegraphics[width=\textwidth]{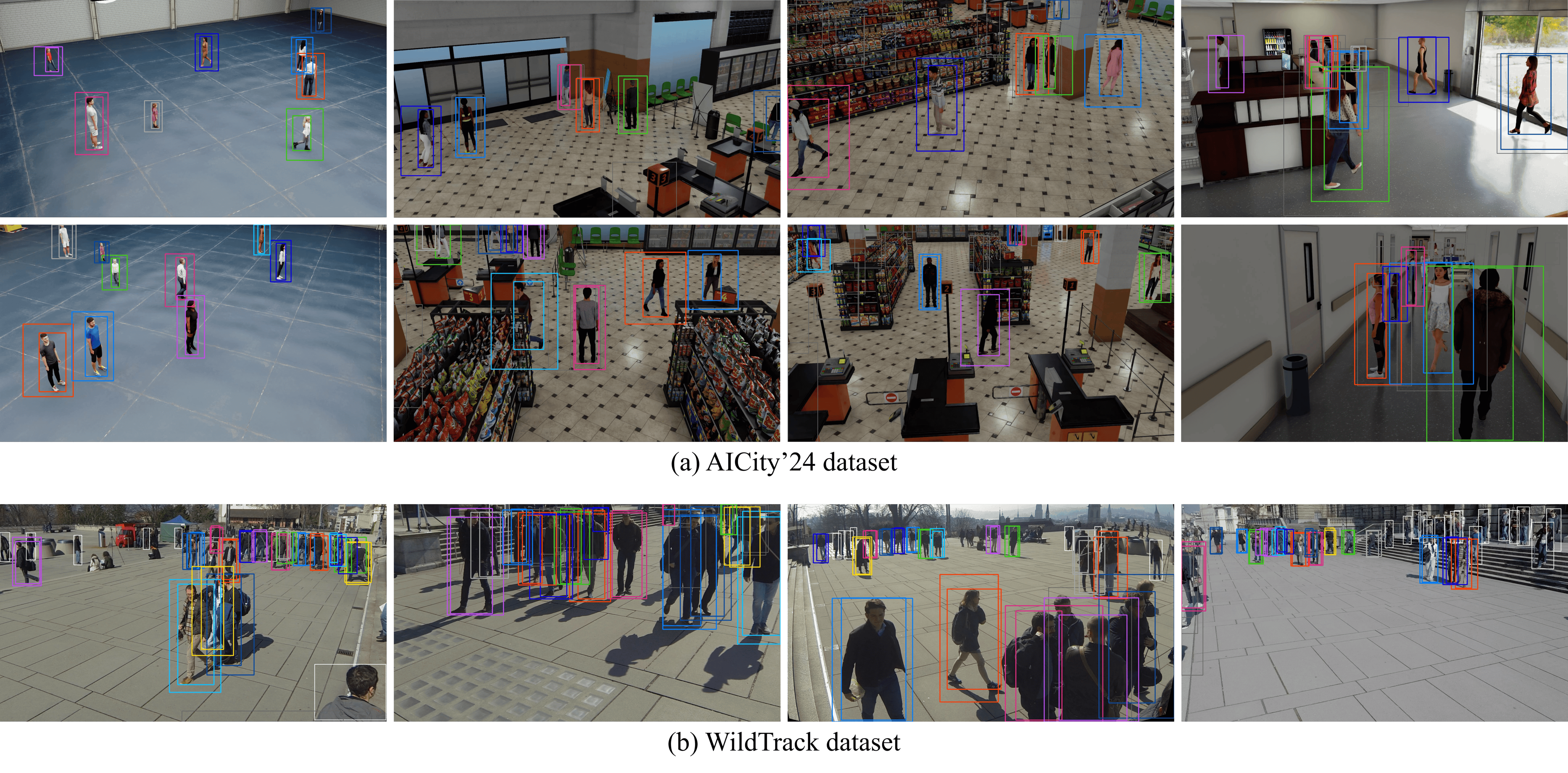}
    \vspace*{-0.5cm}
  \caption{Visualization of 2D-3D detection association results.}
  \label{fig:viz_det_assoc}
\end{figure*}

\subsection{Visualization}
\label{subsec:supp_det_assoc_viz}

We visualized some 2D-3D detection association results on sample frames of the AICity'24 and WildTrack datasets in~\cref{fig:viz_det_assoc}. The associated bounding boxes are in the same color, where the smaller ones are 2D detections and the larger ones are projected 3D detections from BEVFormer. Those 2D detections in white are not associated with any 3D detections.

\subsection{Improvements with Detection Association}
\label{subsec:supp_det_assoc_exp}

We compared the tracking performance of MCBLT with and without the proposed 2D-3D detection association algorithm in~\cref{tab:supp_det_assoc}. The baseline result is based on the ReID features extracted from the large projected 3D bounding boxes shown in~\cref{fig:det_assoc_example}. With the noisy background or other objects included in the image crops, ReID feature quality will be significantly affected. 

\begin{table}[h]
  \centering
  \small
  \begin{tabular}{@{\hspace{1mm}}l|C{0.8cm}C{0.8cm}C{0.8cm}C{0.6cm}C{0.6cm}@{\hspace{2mm}}}
    \toprule
    Method & IDF1 & MOTA & MOTP & MT & ML \\
    \midrule
    Baseline & 63.2 & 73.4 & 93.7 & 24.0 & 4.0 \\
    + det association & 93.4 & 87.5 & 94.3 & 90.2 & 2.4 \\
    \bottomrule
  \end{tabular}
  \caption{A comparison of results on the WildTrack test set with our 2D-3D detection association algorithm.}
  \label{tab:supp_det_assoc}
\end{table}

\section{ReID Feature Quality Analysis}
\label{sec:supp_reid}

We conducted ReID feature quality analysis on both the AICity'24 and WildTrack datasets. For the AICity'24 dataset, we sampled 500 characters with their 2D bounding boxes and object IDs from the ground truth across all scenes and cameras from the test set. The total object image crop count is 40,000. We filtered out 2D bounding boxes that are smaller than 5,000 pixels, as well as those whose aspect ratio (\ie, width / height) is less than 0.15. 
Similarly, for the Wildtrack dataset, we sampled 330 characters from the sequence and applied the same filters bringing the total object crop count to 41,284.

\begin{table}[h]
  \centering
  \small
  \begin{tabular}{l|cccc}
    \toprule
    Dataset & Rank-1 & Rank-5 & Rank-10 & mAP \\
    \midrule
    AICity'24 & 95.02 & 97.44 & 98.08 & 73.85 \\
    WildTrack & 77.18 & 84.49 & 87.97 & 63.11 \\
    \bottomrule
  \end{tabular}
  \caption{Evaluation on our ReID feature quality.}
  \label{tab:supp_reid_eval}
\end{table}

We evaluated the ReID feature quality by the mean average precision (mAP), rank-1, rank-5, and rank-10 accuracies. The evaluation results are shown in~\cref{tab:supp_reid_eval}.
We found that the feature quality on the WildTrack dataset is worse than that on the AICity'24 dataset. This is because i) WildTrack is a real-world dataset with more noises and diverse illuminations from different camera views; ii) 2D bounding box annotations are not as accurate as the synthetic AICity'24 dataset. Nevertheless, our MCBLT achieved impressive results on WildTrack based on these ReID features.

\section{Model Time Complexity Analysis}

Although MTMC detection and tracking tasks do not usually require real-time performance and are tolerant to time delays, we record the running time of the proposed MCBLT pipeline in~\cref{tab:time_complexity} to provide a rough impression of the complexity of the model. The model inference is conducted on one single NVIDIA A100 GPU, with 10 cameras in the scene. Our method achieves around 1.5 FPS end-to-end before any further model optimization. The 2D detection, ReID, and tracking models are very efficient and can operate in parallel with BEVFormer so that their running time is negligible.

\begin{table}[h]
    \centering
    \begin{tabular}{c|c|c}
        \hline
        \multicolumn{2}{c|}{Detection} & Tracking                  \\
        \hline
        \multicolumn{2}{c|}{{\scriptsize\textcolor{gray}{\textsc{BEVFormer}}} \ \ 1.6 FPS } & \multirow{2}{*}{\begin{tabular}[c]{@{}c@{}}{\scriptsize\textcolor{gray}{\textsc{SUSHI}}} \\ 452.7 FPS \end{tabular}} \\
        \cline{1-2}
        {\scriptsize\textcolor{gray}{\textsc{DINO}}} \ 65.0 FPS & {\scriptsize\textcolor{gray}{\textsc{SOLIDER}}} \ 58.9 FPS &   \\
        \hline
    \end{tabular}
    \caption{MCBLT model efficiency analysis.}
    \label{tab:time_complexity}
\end{table}

\section{Overall Visualization}
\label{sec:supp_viz}

We also visualized MTMC detection and tracking results of MCBLT on the AICity'24 and WildTrack datasets. Please find the demos in the attachment.

\end{document}